\documentclass[5p,times,twocolumn]{elsarticle}
\usepackage{lineno,hyperref}

\usepackage{tikz}
\usepackage{pgfplotstable}
\usepackage{pgfplots}
\usepackage{multirow}
\usepackage{float}
\usepackage{textcomp}
\usepackage{amsmath,amsfonts,amssymb,bbm}	
\usepackage{multicol}
\usepackage{subfigure}
\usepackage{amsmath}
\usepackage{amssymb}
\usepackage{autobreak}
\usepackage{algorithm}
\usepackage{algorithmic}

\usepackage{siunitx}
\usepackage{graphicx} 
\usepackage{lscape} 
\usepackage{booktabs} 
\usepackage{footmisc} 
\usepackage{footnote} 
\makesavenoteenv{tabular} 
\makesavenoteenv{table} 

\makeatletter
\def\ps@pprintTitle{%
  \let\@oddhead\@empty
  \let\@evenhead\@empty
  \let\@oddfoot\@empty
  \let\@evenfoot\@oddfoot
}
\makeatother

\usepackage[section]{placeins}

\modulolinenumbers[5]

\begin{document}

\begin{frontmatter}

\title{Data Augmentation For Medical MR Image Using Generative Adversarial Networks}


\author{Panjian Huang\fnref{myfootnote}}
\fntext[myfootnote]{This work is done when Panjian Huang was an intern at WATRIX.AI.}
\author{Xu Liu\corref{mycorrespondingauthor}}
\author{Yongzhen Huang}
\address{\{panjian.huang, xu.liu, hyz\}@watrix.ai}

\address{WATRIX.AI}

\begin{abstract}
Computer-assisted diagnosis (CAD) based on deep learning has become a crucial diagnostic technology in the medical industry, effectively improving diagnosis accuracy. However, the scarcity of brain tumor Magnetic Resonance (MR) image datasets causes the low performance of deep learning algorithms. The distribution of transformed images generated by traditional data augmentation (DA) intrinsically resembles the original ones, resulting in a limited performance in terms of generalization ability. This work improves Progressive Growing of GANs with a structural similarity loss function (PGGAN-SSIM) to solve image blurriness problems and model collapse. We also explore other GAN-based data augmentation to demonstrate the effectiveness of the proposed model. Our results show that PGGAN-SSIM successfully generates 256x256 realistic brain tumor MR images which fill the real image distribution uncovered by the original dataset. Furthermore, PGGAN-SSIM exceeds other GAN-based methods, achieving promising performance improvement in Frechet Inception Distance (FID) and Multi-scale Structural Similarity (MS-SSIM).
\end{abstract}


\end{frontmatter}


\section{Introduction}
Convolutional Neural Networks (CNNs) have achieved remarkable success in medical image analysis, such as brain tumor MRI segmentation~\cite{havaei2017brain,ning2021smu}. However, training a high-performance medical analysis algorithm requires large-scale artificially labeled training data sets as small-scale image datasets with low diversity will decrease the accuracy and generalization of the algorithm. Data augmentation (DA) technology becomes the primary and vital link in combining artificial intelligence and medical fields~\cite{guo2020anatomic,kazeminia2020gans}. Although traditional data augmentation techniques (such as flip, translation, rotation, cropping) can alleviate the small number of medical samples, they produce highly relevant training data without new image information~\cite{nie2017medical}. 

Recently, generative adversarial networks (GANs)~\cite{goodfellow2014generative} have shown excellent performances in the visual image generation task, such as Wasserstein GAN with Gradient Penalty (WGAN-GP)~\cite{gulrajani2017improved} for stabilizing training process, StyleGAN~\cite{karras2019style} for controlling image attributes. Some previous works have shown the potential of GAN-based data augmentation in medical image generation~\cite{kim2021synthesis}. Han et al.~\cite{han2018gan} proposed WGAN and DCGAN to generate four modalities of brain MR image avoiding artifacts, evenly an expert physician was unable to accurately distinguish the generated images from the real samples in the Visual Turing Test. However, their methods only generate 64 x 64 and 128 x 128 low resolution brain tumor MR images, while most CNNs architecture adopt around 256 x 256 input size (e.g., ResNet-50: 224 x 224~\cite{he2016deep}) to acquire more representation information from images~\cite{han2018infinite}. Kwon et al.~\cite{kwon2019generation} improved $\alpha$-GAN with gradient penalty ($\alpha$-GAN-GP) to generate realistic 3D brain tumor MR images, combining Variational Auto-Encoder (VAE)~\cite{kingma2013auto} and GAN with an additional discriminator to solve mode collapse and image blurriness problems. However, the complexity of 3D space and deconvolutional layers in their model leads to inconsistency of 3D MR image sequences and checkerboard artifacts respectively~\cite{shi2016real,zhao2020smore}. Han et al.~\cite{han2018infinite} exploited Progressive Growing GANs (PGGANs) to generate 256 x 256 high-resolution brain tumor MR images. Although the generated samples by PGGANs improved the accuracy of tumor detection using ResNet-50, their training data is consisted of different slices in brain tumor MR images, resizing to 256 x 256 resolution. That means that the training data is composed of different feature spaces, which may cause PGGAN to generate failed samples. Besides, due to the low prevalence of brain tumors and the privacy of patients involved in medical data, it is challenging to obtain brain tumor MR images in the medical field~\cite{nie2017medical}.

Therefore, a well-suited GAN model to generate 256 x 256 resolution realistic brain tumor MR images with diversity is essential. The main challenges are as follows: (i) GAN model needs to overcome the unstable training with high-resolution inputs and mode collapse due to an inappropriate loss function; (ii) GAN model on small-scale training data is difficult to synthesize realistic brain MR images from a random vector. To address the mentioned challenges in medical image generation, our main contributions are:
\begin{itemize}
\item \textbf{GAN Selection:} This research explores two completely different GAN models ($\alpha$-GAN-GP, PGGAN) to select the suitable architecture to overcome the unstable training and generate realistic brain MR images. Additionally, we also propose PGGAN with the SSIM function~\cite{wang2004image} (PGGAN-SSIM) to improve the mode collapse.

\item \textbf{Brain Tumor MRI Generation:} This research shows that PGGAN-based with new loss function is able to generate promising results, exceeding existing GAN-based methods both in FID~\cite{mathiasen2020fast} and MS-SSIM~\cite{wang2003multiscale}. The synthesized 256 x 256 brain tumor MR images are realistic and diverse, possibly improving the accuracy of medical image detection.
\end{itemize}

\section{Background}
\subsection{Gliomas MRI}
Although glioma is a complex tumor composed of multiple tumor tissues (such as necrotic core, active tumor edge, edema tissue), it can be scanned by multiple Magnetic Resonance Imaging (MRI) sequences~\cite{basu2011fundamentals27}. MRI has the characteristics of multi-modal imaging, and different imaging sequences can be obtained through different imaging techniques, including four modalities: T1-weighted (T1), contrast-enhanced T1-weighted (T1c), T2-weighted (T2), and Fluid Attenuation Inversion Recovery (FLAIR) ~\cite{bauer2013survey28,valentino1991volume29}.
\begin{figure}[!tbp]
	\centering  
	\subfigure[T1]{
		\label{T1}
		\includegraphics[width=0.23\linewidth,angle=90]{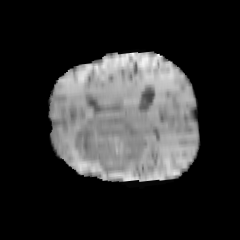}}
	\subfigure[T1ce]{
		\label{T1ce}
		\includegraphics[width=0.23\linewidth,angle=90]{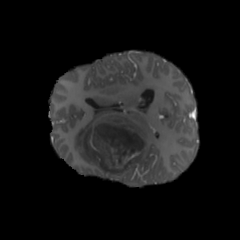}}
		\subfigure[T2]{
		\label{T2}
		\includegraphics[width=0.23\linewidth,angle=90]{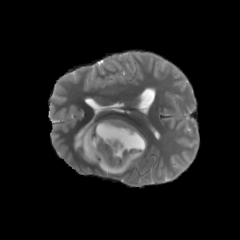}}
		\subfigure[FLAIR]{
		\label{FLAIR}
		\includegraphics[width=0.23\linewidth,angle=90]{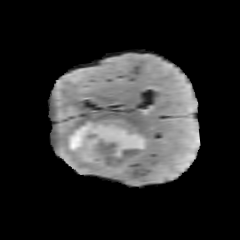}}	
	\caption{Four modalities of Gliomas.}
	\label{Fig.main}
\end{figure}\\
It can be seen from Figure~\ref{Fig.main} that the same organization is displayed differently in different modalities~\cite{valentino1991volume29}. The brain tissue is relatively constant in the T1 and T2. The edema area is dark in the T1 modal, while the white shadow area is formed on the T2. The T1 pays more attention to the anatomical details of each tissue; The T2 can highlight the lesion area more clearly and retain a high signal to the cerebrospinal fluid~\cite{yu2020sample}. FLAIR uses water suppression technology for imaging. Free water is a low signal (edema state), and bound water is a high signal (non-edema area)~\cite{valentino1991volume29}. The FLAIR sequence can clearly show the lesions near the ventricle or near the cortex~\cite{won2021development}. That is useful for observation in the diagnosis of diseases such as gliomas. Therefore, this work mainly focuses on the FLAIR sequence generation of gliomas images.
\subsection{Generative Adversarial Networks}
Since GAN was proposed in 2014 by Goodfellow et al.~\cite{goodfellow2014generative}, it has achieved dramatic success in computer vision. As various variants of GANs are presented, the researches of applying GANs to the medical image field have been further developed, like medical image-to-image translation, medical image resolution, and medical image detection~\cite{shen2017deep,tuysuzoglu2018deep,yi2019generative}.

\begin{figure}[!tbp]
\centering
\includegraphics[width=0.5\linewidth]{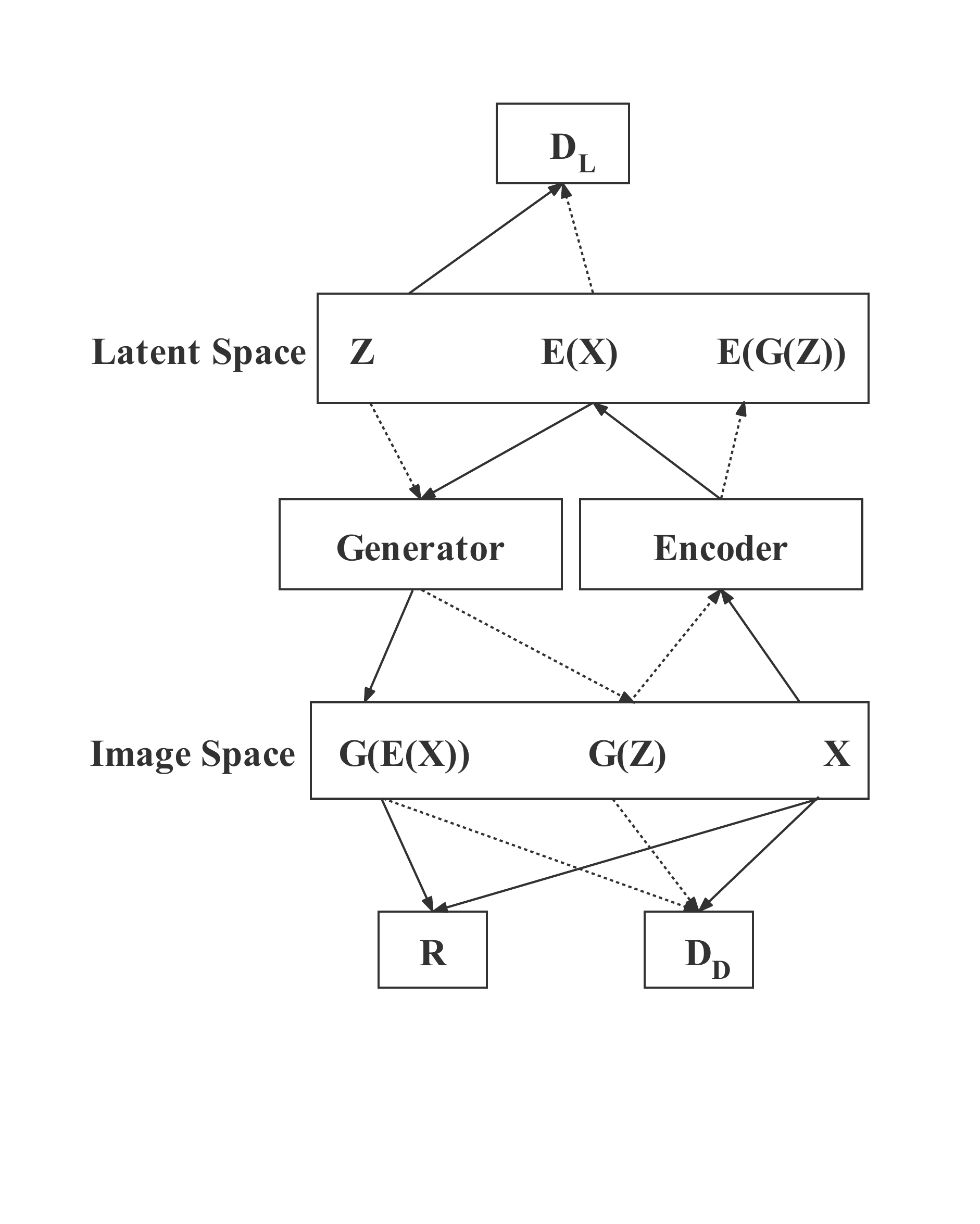} 
\caption{Structure of $\alpha$-GAN-GP.} 
\label{Fig.main2}
\end{figure}

Kwon et al. proposed $\alpha$-GAN-GP that consists of a generator, an encoder, a discriminator, and an additional discriminator L~\cite{rosca2017variational10}, combining VAE and GAN to map random noises to a gliomas MRI distribution successfully. The special structure of $\alpha$-GAN-GP is shown as Figure~\ref{Fig.main2}~\cite{rosca2017variational10}. VAE attempts to compress discrete pixels in the real training image into a low dimensional continuous space and then reconstruct it back to the original space through encoding~\cite{rosca2017variational10}. KL divergence is introduced in the encoder and decoder to measure the difference between the distribution of latent variables and the Gaussian unit distribution, forcing the generated image distribution to approach the real image distribution continuously~\cite{rosca2017variational10}. However, the farther away the generated image by VAE is from the center point, the more blurred it is. Furthermore, the generated image by VAE lacks a detailed description, and it is not easy to generate high-quality images with precise details. Therefore, the combination idea of original $\alpha$-GAN is: (i) although the VAE will generate blurry images, it can avoid the mode collapse problem since all real samples will be encouraged to be reconstructed by an autoencoder; (ii) GAN does not need further constraints on the model, generating more detailed and sharper images than VAE, but GAN will suffer from mode collapse~\cite{rosca2017variational10}. However, although $\alpha$-GAN can overcome the image blurriness and mode collapse, it is still difficult to train since when the real data distribution and the generated data distribution do not have overlapping areas or can ignore overlapping areas, the JS divergence is usually a constant, which will cause the generator gradient to disappear~\cite{wolterink2018generative}. Furthermore, the gradient exploding or vanishing may still occur. Therefore, Kwon et al.~\cite{kwon2019generation} exploited the gradient penalty (GP) to constraint the training process of $\alpha$-GAN, called $\alpha$-GAN-GP.

\begin{figure}[!ht]
\centering
\includegraphics[width=1.0\linewidth]{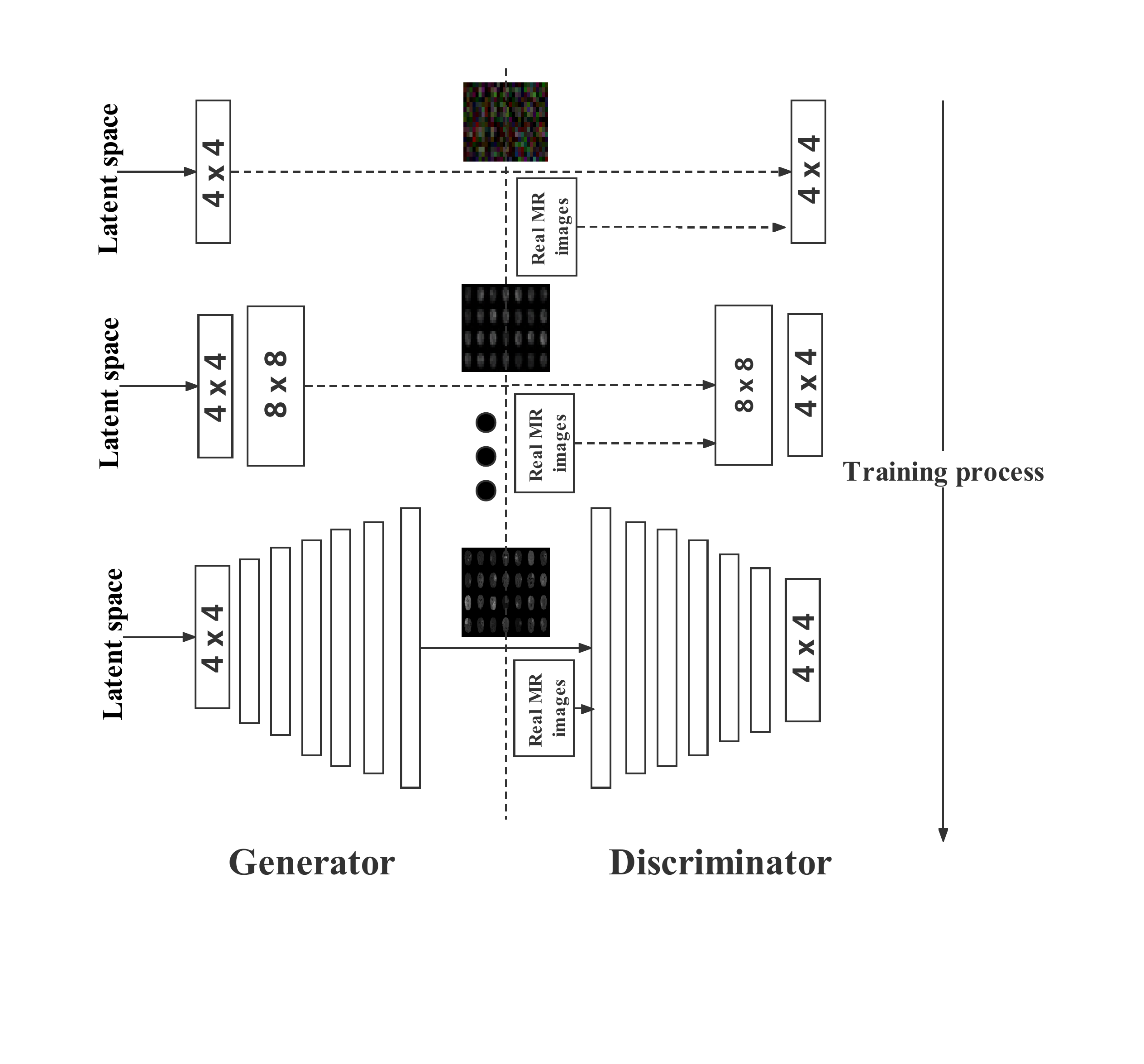} 
\caption{PGGAN training process for 256 x 256 MR image generation.} 
\label{Fig.main3}
\end{figure}

Han et al.~\cite{han2018infinite} exploited PGGAN architecture to synthesize original-sized 256 x 256 realistic brain MR images, improving the accuracy of tumor detection. PGGAN proposed by Karras et al.~\cite{karras2017progressive} is firstly able to generate 1024 x 1024 realistic images as it uses a novel training strategy with a progressively growing generator and discriminator: starting from low resolution, newly added layers capture fine-grained details as training progresses. As Figure~\ref{Fig.main3} shows, they successfully adopt PGGAN to generate realistic 256 x 256 brain MR images.

In our research, we address the quality and diversity of synthesized samples through FID and MS-SSIM. Unlike previous works, We select 64th brain tumor FLAIR images in training data, discarding the remaining sequence images containing redundant and interference information. Two basic GAN architectures are explored for training stability and model collapse. Additionally, we illustrate the effect of the SSIM loss function to improve the diversity of the generated distribution in GAN-based data augmentation. We creatively improve PGGAN with an SSIM loss function to generate 256 x 256 realistic gliomas FLAIR sequence images with diversity on a small-scale training dataset.

\section{Method}
This research addresses two questions: GAN selection and Realistic brain tumor generation with high resolution. Therefore, we compare PGGAN-SSIM with two baseline models ($\alpha$-GAN-GP, PGGAN), to find a suitable GAN between them for medical image generation. In this section, we describe GAN-based methods in detail.
~\\

\noindent\textbf{$\alpha$-GAN-GP} There are four networks: encoder, generator, discriminator, and discriminator L to be trained. For the encoder, the input of the encoder is real data x, and then it will output a latent vector $\hat{z}$, which will input to the discriminator L. The purpose of the encoder is to deceive the discriminator L into treating the $\hat{z}$ as real. For the generator, the inputs of the generator are random noise $z$ and $\hat{z}$, and then it will output the generated image and the reconstructed image from the two vectors. The generator aims to fool the discriminator to treat the two output images as real and reduce the difference between the real image and the reconstructed image~\cite{goodfellow2014generative}. For the discriminator, the inputs are real images, generated images, and reconstructed images. The discriminator aims to classify these images, treating the generated images and the reconstructed images as fake~\cite{goodfellow2014generative}. The inputs of discriminator L are $z$ and $\hat{z}$. The purpose of the discriminator L is to classify the two classes of vectors, treating the $\hat{z}$ as real, so that the distribution $\hat{z}$ can be as close as possible to $z$ normal distribution. The total loss is below:

\begin{figure}[!h]
    \centering
\begin{scriptsize}
\begin{equation}
\left\{
\begin{array}{lr}
Generator\quad Loss  = - \mathop{\mathbb{E}}\limits_{x\sim \mathbb{P}_{r}}[D_{D}(G(E(x)))] - \mathop{\mathbb{E}}\limits_{z\sim \mathbb{P}_{z}}[D_{D}(G(z))]+\lambda_{1}\|x-G(E(x))\|_{L1}\\
Encoder\quad Loss = - \mathop{\mathbb{E}}\limits_{x\sim \mathbb{P}_{r}}[D_{L}(E(x))]\\
Discriminator\quad Loss = \mathop{\mathbb{E}}\limits_{x\sim \mathbb{P}_{r}}[D_{D}(G(E(x)))] + \mathop{\mathbb{E}}\limits_{z\sim \mathbb{P}_{z}}[D_{D}(G(z))]\\
\quad \quad \quad \quad \quad \quad \quad \quad \quad \quad - 2\mathop{\mathbb{E}}\limits_{x\sim \mathbb{P}_{r}}[D_{D}(x)] +  \lambda_{2}L_{GP-D_{D}}\\
Discrimiantor\quad L\quad Loss = \mathop{\mathbb{E}}\limits_{x\sim \mathbb{P}_{r}}[D_{L}(E(x))] - \mathop{\mathbb{E}}\limits_{z\sim \mathbb{P}_{z}}[D_{L}(z)] + \lambda_{2}L_{GP-D_{L}}\\
\end{array}
\right.
\end{equation}
\end{scriptsize}
\end{figure}

\noindent Where $D_{D}$ denotes the discriminator of the data space; $D_{L}$ represents the discriminator of the latent space; E is the encoder; G denotes the generator. The discriminator loss of $D_{D}$ contains the sum of two distance metrics since the discriminator treats the reconstructed images $G(E(x))$ and randomly generated images $G(z)$ as fake~\cite{kwon2019generation}. The generator loss uses the L1 distance between the reconstructed images $G(E(x))$ and real images $x$ as the reconstruction term. The encoder loss has an identical generator loss form of WGAN-GP. Similarly, the discriminator $L$ loss has the identical discriminator loss form of WGAN-GP except for the input, where the input is a latent variable $z_{r}$ or $z_{e}$ instead of images. The gradient penalty~\cite{arjovsky2017towards} terms $L_{GP-D_{D}}$ and $L_{GP-D_{L}}$ are added to $D_{D}$ and $D_{L}$. The gradient penalty in discriminator is described as $\mathop{\mathbb{E}}\limits_{\hat{x}\sim\mathbb{P}_{\hat{x}}}[(\|\nabla_{\hat{x}}D_{D}(\hat{x})\|_{2}-1)^2]$, where $\hat{x}$ is any point sampled between real and generated samples. For the parameters $\lambda_{1}$ and $\lambda_{2}$, they use the fixed value of 10~\cite{kwon2019generation}. In discriminator L, the gradient penalty is described as $\mathop{\mathbb{E}}\limits_{\hat{z}\sim\mathbb{P}_{\hat{z}}}[(\|\nabla_{\hat{z}}D_{L}(\hat{z})\|_{2}-1)^2]$, where $\hat{z}$ is any point sampled between encoding and random noise samples.
~\\

\noindent\textbf{PGGAN} There are mainly three improvements that enable PGGAN to generate high-quality 1024 x 1024 high-resolution images:
\begin{itemize}
\item \textbf{Progressive Growing Training Strategy:} During training, new blocks of convolutional layers are systematically added to both the generator model and the discriminator models~\cite{karras2017progressive}. The incremental addition of the layers allows the models to learn coarse-level detail effectively and later learn ever finer detail, both on the generator and discriminator side. Attentionally, When new layers are added to the networks, PGGAN fades current layers in smoothly. That avoids sudden shocks to the already well-trained, smaller-resolution layers. Further, all layers remain trainable during the training process, including existing layers when new layers are added.
\item \textbf{Increasing Variation Using Minibatch Standard Deviation:} The minibatch layer in discriminator measures the diversity within a minibath, which enable generator to escape mode collapse~\cite{karras2017progressive}.
\item \textbf{Equalized Learning Rate and Pixelwise Feature Vector Normalization}~\cite{karras2017progressive}
\end{itemize}

\begin{figure}[!tbp]
\centering
\includegraphics[width=1.0\linewidth]{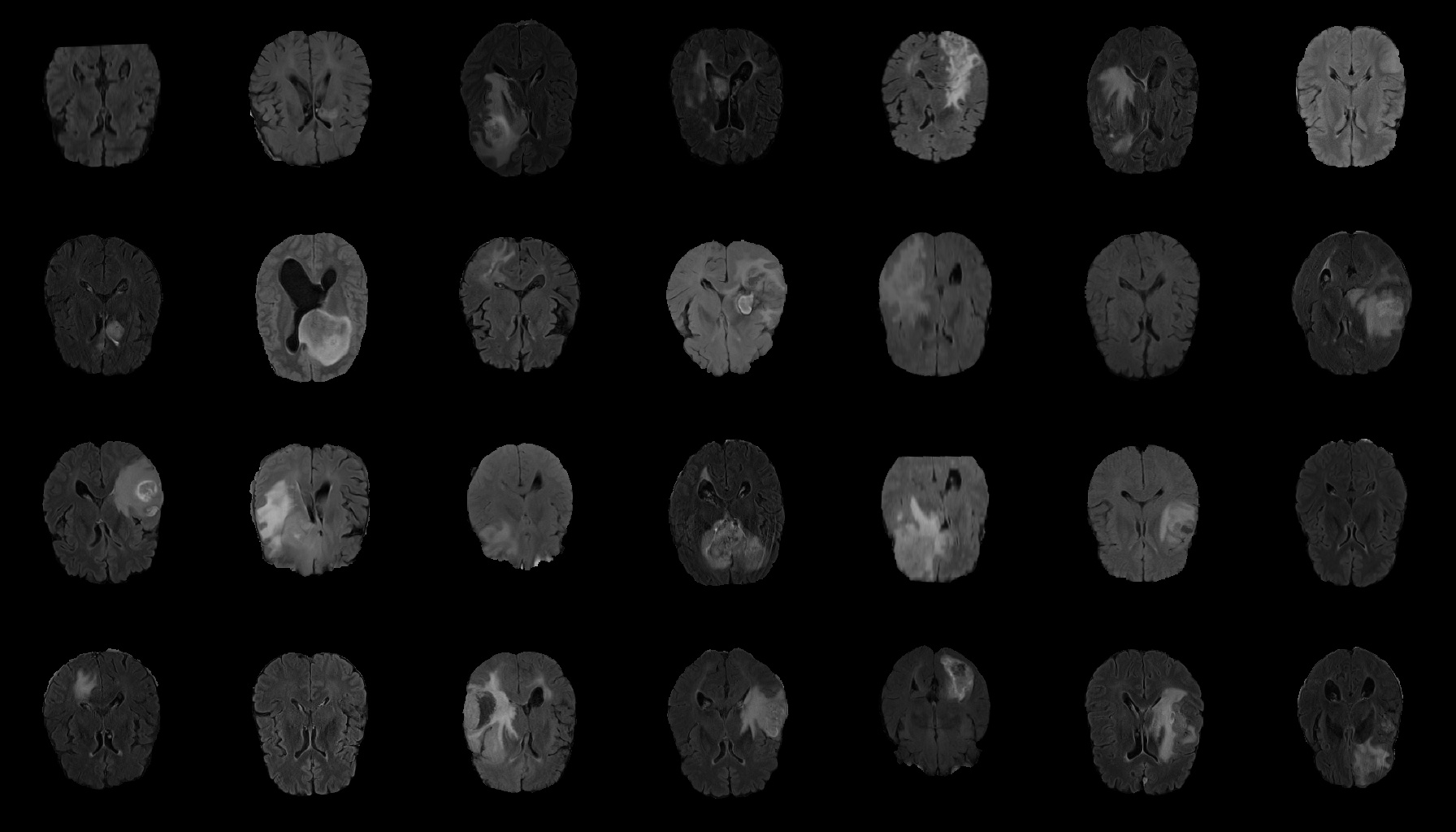} 
\caption{Real FLAIR samples in AXI for GAN training.} 
\label{Fig.main4}
\end{figure}

PGGAN utilizes WGAN-GP loss function as follows:

\begin{figure}[!h]
    \centering
\begin{scriptsize}
\begin{equation}
\left\{
\begin{array}{lr}
Generator\quad Loss = - \mathop{\mathbb{E}}\limits_{x\sim \mathbb{P}_{r}}[D(x)] \\
Discriminator\quad Loss = \mathop{\mathbb{E}}\limits_{\tilde{x}\sim\mathbb{P}_{g}}[D(\tilde{x})] + \lambda\mathop{\mathbb{E}}\limits_{\hat{x}\sim\mathbb{P}_{\hat{x}}}[(\|\nabla_{\hat{x}}D(\hat{x})\|_{2}-1)^2]\\
\end{array}
\right.
\end{equation}
\end{scriptsize}
\end{figure}

\noindent Where $\mathbb{P}_{\hat{x}}$ represents the sample distribution randomly along with the space between the data distribution $\mathbb{P}_{r}$ and the generator distribution $\mathbb{P}_{g}$~\cite{gulrajani2017improved}.
~\\

\noindent\textbf{PGGAN-SSIM} There are no differences between PGGAN-SSIM and original PGGAN architectures. However, due to the small-scale training dataset and the size limitation of the minibatch layer in the discriminator, mode collapse possibly happens in the training process. Therefore, we add the SSIM loss function to the generator, forcing it to synthesize images with more diversity. The total loss is as follows:
\begin{figure}[!h]
    \centering
\begin{scriptsize}
\begin{equation}
\left\{
\begin{array}{lr}
Generator\quad Loss = - \mathop{\mathbb{E}}\limits_{x\sim \mathbb{P}_{r}}[D(x)] + L_{SSIM} \\
Discriminator\quad Loss = \mathop{\mathbb{E}}\limits_{\tilde{x}\sim\mathbb{P}_{g}}[D(\tilde{x})] + \lambda\mathop{\mathbb{E}}\limits_{\hat{x}\sim\mathbb{P}_{\hat{x}}}[(\|\nabla_{\hat{x}}D(\hat{x})\|_{2}-1)^2]\\
\end{array}
\right.
\end{equation}
\end{scriptsize}
\end{figure}

\section{Experiments}
This research exploits the BRATS 2019 training dataset containing 259 High-Grade Gliomas (HGG) cases. We select the 64th slice among the whole 155 slices as the initial/final slices will convey a negligible amount of information and affect the training~\cite{dar2019image}. Therefore, the training dataset is 259 FLAIR brain axial (AXI) MR images, zero-padded to 256 x 256 from original-sized 240 x 240 pixels for better GAN training. Figure~\ref{Fig.main4} shows real FLAIR samples.

\begin{table}[!tbp]
	\centering
	\begin{tabular}{c|c|c}
		\toprule
		Model& FID& MS-SSIM\\
		\midrule
		$\alpha$-GAN-GP (z256)& 152.71& 0.7562\\
		$\alpha$-GAN-GP (z512)& 127.15& 0.7206\\
		\midrule
		PGGAN (z512)& \textbf{32.272}& 0.6651\\
		PGGAN (z1024)& 40.900& 0.6685\\
		\midrule
		PGGAN-SSIM (z512)& 33.266& \textbf{0.6617}\\
		PGGAN-SSIM (z1024)& 39.423& 0.6619\\
		\midrule
		Real& \--& 0.6627\\
		\bottomrule
	\end{tabular}
	\caption{Quantitative results over HGG with FLAIR in Brats 2019 training dataset with models: $\alpha$-GAN-GP, PGGAN and PGGAN-SSIM.}\label{quantitative}
\end{table}

\begin{figure}[!tbp]
	\centering  
	\subfigure[$\alpha$-GAN-GP]{
		\label{Fig.sub.2}
		\begin{minipage}[b]{0.15\textwidth}
		\includegraphics[width=1\textwidth]{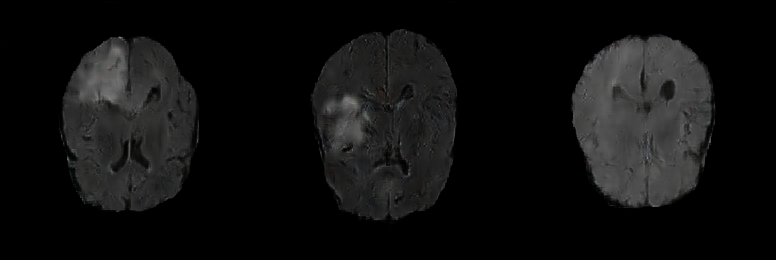}\vspace{-1mm}
		\includegraphics[width=1\textwidth]{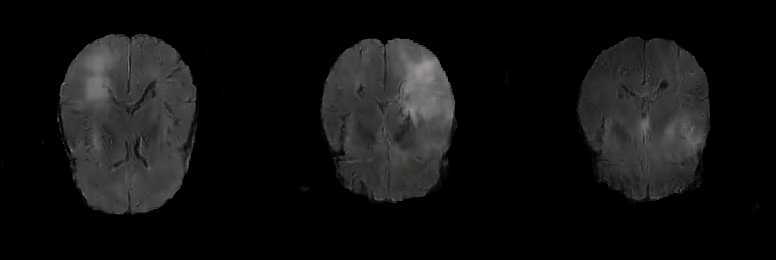}\vspace{-1mm}
		\includegraphics[width=1\textwidth]{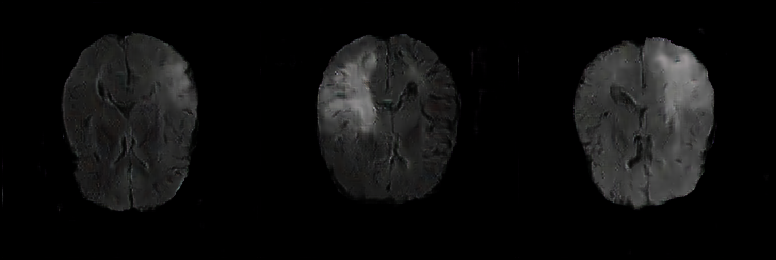}\vspace{-1mm}
		\end{minipage}}\hspace{-1mm}
	\subfigure[PGGAN]{

		\label{Fig.sub.2}
		\begin{minipage}[b]{0.15\textwidth}
		\includegraphics[width=1\textwidth]{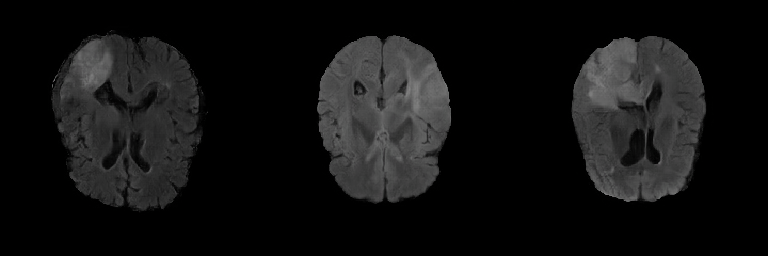}\vspace{-1mm}
		\includegraphics[width=1\textwidth]{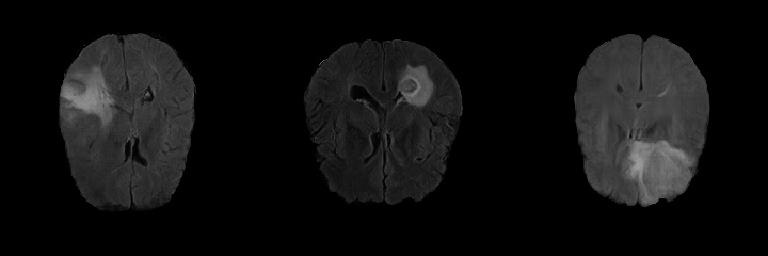}\vspace{-1mm}
		\includegraphics[width=1\textwidth]{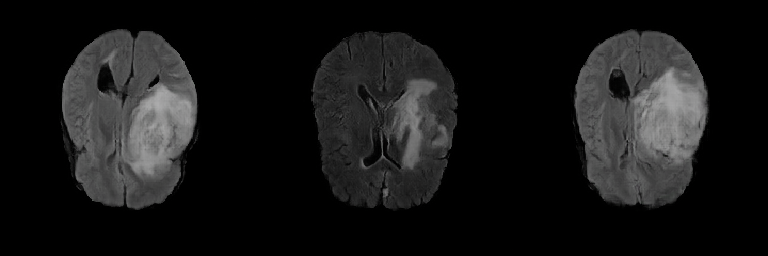}\vspace{-1mm}
		\end{minipage}}\hspace{-1mm}
	\subfigure[PGGAN-SSIM]{

		\label{Fig.sub.2}
		\begin{minipage}[b]{0.15\textwidth}
		\includegraphics[width=1\textwidth]{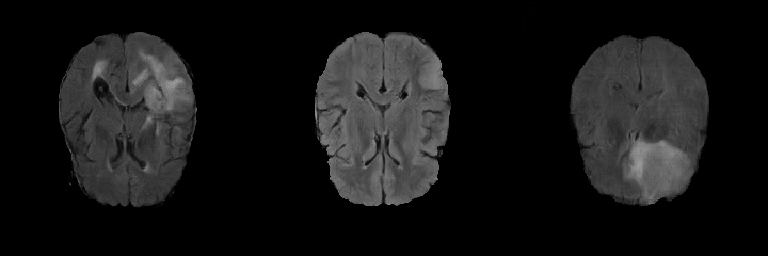}\vspace{-1mm}
		\includegraphics[width=1\textwidth]{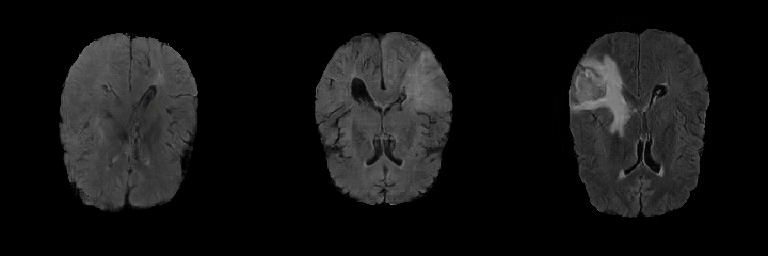}\vspace{-1mm}
		\includegraphics[width=1\textwidth]{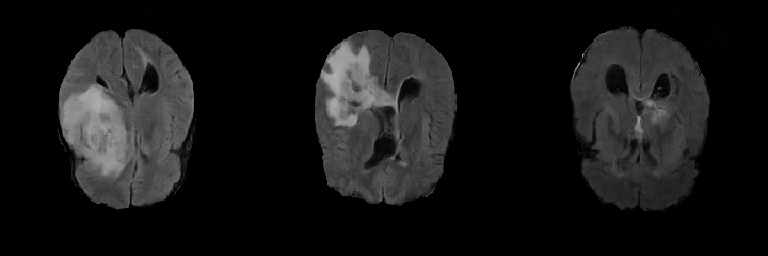}\vspace{-1mm}
		\end{minipage}}\hspace{-1mm}
	\caption{Generated samples of FLAIR brain AXI MR images.}
	\label{compare1}
\end{figure}

Our experiments are conducted on 8 NVIDIA 3090 24GB GPU. All models are trained for 12000 iterations with a batch size of 32, the learning rate of 0.001 for Adam, and the coefficient of gradient penalty of 10.0. For PGGAN-SSIM, the coefficient of SSIM is 10.0.

The z256, z512, and z1024 represent the different models with different input vector sizes of 256, 512, and 1024 dimensions.

\noindent\textbf{Generated Images} Figure~\ref{compare1} shows the generated brain tumor FLAIR samples from $\alpha$-GAN-GP, PGGAN, and PGGAN-SSIM in AXI. The generated samples from all models look realistic and are close to the real images, where the white areas (Gliomas area) are concentrated and unified, indicating the training stability with gradient penalty. However, the samples from $\alpha$-GAN-GP are blurry, and the detailed brain features disappear since the loss of image information becomes serious when the image resolution increases in $\alpha$-GAN-GP. The generated samples from PGGAN and PGGAN-SSIM have clear and realistic brain features (FLAIR texture and tumor appearance) with diversity, which illustrates that PGGAN architecture is a well-suited model to generate brain MR images.

\noindent\textbf{Quantitative Analysis} We measure FID to calculate the distribution distance between real samples and generated data. If the FID score is lower, the generated samples by the model are closer to the real sample, indicating better model performance. We select 10000 pairs of test data to calculate FID scores for each model. Table~\ref{quantitative} illustrates that PGGAN with the input size of 512 has the lowest FID score, indicating the generated sample by PGGAN looks most realistic.

We also investigate MS-SSIM to evaluate the diversity of the generated distribution. The MS-SSIM score is obtained by calculating the average from 2000 sample pairs. The similarity between the generated samples by $\alpha$-GAN-GP is significant, suffering from mode collapse. The PGGAN and PGGAN-SSIM have similar MS-SSIM scores to real data, evenly PGGAN-SSIM outperforms the real data, which means that the generated samples have more diversity. Therefore, the structure of PGGAN can effectively avoid mode collapse, and the SSIM loss function can further improve the performance.

\noindent\textbf{Ablation Study} We also explore the effect of different input sizes and SSIM loss function. It can be seen from Table~\ref{quantitative} that when the input size changes from 512 to 1024, the FID and MS-SSIM scores of all models become higher. The best results of this experiment are obtained by using a moderately sizeable latent vector size of 512. There is a trade-off balance between w/o SSIM loss function. When adding an SSIM loss function to the generator in PGGAN, the MS-SSIM scores are improved below the real data while the FID scores are a bit degraded. Since our objective is data augmentation that needs more samples different from real data, we suggest adding the SSIM loss function to improve the sample diversity.

\section{Conclusion}
In this work, we explore the GAN-based data augmentation to generate realistic 256 x 256 brain tumor MR images with diversity on a small amount of training data. We investigate two baseline models ($\alpha$-GAN-GP, PGGAN) to overcome the mode collapse and training instability. We also propose PGGAN with an SSIM loss function (PGGAN-SSIM) to improve the diversity of generated samples. The results show that PGGAN-SSIM can successfully generate high-quality 256 x 256 brain tumor FLAIR images with more diversity, exceeding performances of other GAN-based models both in FID and MS-SSIM.

In future work, we plan to show the improvement in tumor detection with the generated samples. Additionally, We will explore more excellent GAN architectures in the medical image generation task.

\bibliography{main}

\end{document}